# A Technique for Classifying Static Gestures Using UWB Radar

Abhishek Sebastian [1][0000-0002-3421-1450] and Pragna R [1][0000-0003-0827-5896]

[1] Abhira, Department of Applied AI, Chennai

**Abstract.** Our paper presents a robust framework for UWB-based static gesture recognition, leveraging proprietary UWB radar sensor technology. Extensive data collection efforts were undertaken to compile datasets containing five commonly used gestures. Our approach involves a comprehensive data pre-processing pipeline that encompasses outlier handling, aspect ratio-preserving resizing, and false-color image transformation. Both CNN and MobileNet models were trained on the processed images. Remarkably, our best-performing model achieved an accuracy of 96.78%. Additionally, we developed a user-friendly GUI framework to assess the model's system resource usage and processing times, which revealed low memory utilization and real-time task completion in under one second. This research marks a significant step towards enhancing static gesture recognition using UWB technology, promising practical applications in various domains.

## 1      Introduction

Human-Computer Interaction (HCI) has undergone a remarkable evolution over the years, transforming the way we engage with technology in our daily lives [9]. From the early days of punch cards and command-line interfaces to modern touchscreens and voice-activated assistants, HCI has continually strived to make technology more accessible, intuitive, and responsive to human needs [10,11]. In this dynamic landscape, the integration of gesture recognition has emerged as a compelling avenue, promising to elevate HCI to new heights by enabling natural and non-intrusive interactions between humans and machines.

Gestures, the natural language of human expression, have long been a fundamental means of communication [12]. As technology becomes increasingly integrated into our surroundings, the use of gestures as an HCI mechanism offers several distinct advantages. Firstly, it allows for interaction without the physical constraints of traditional input devices, offering a more fluid and liberating experience. Secondly, gestures can bridge language barriers, making technology more inclusive and accessible to a global audience [13]. Lastly, and perhaps most significantly, gesture-based HCI has the potential to enhance the overall user experience by making interactions more intuitive and engaging.

Within the realm of gesture recognition, two primary categories emerge static and dynamic gestures [14]. Static gestures involve specific hand configurations or poses that convey meaning without any substantial movement. These gestures are powerful tools for discrete commands and symbol-based communication. In contrast, dynamic gestures involve fluid hand movements that convey meaning through motion and trajectory. These gestures are ideal for conveying more complex instructions or mimicking real-world actions.

While both static and dynamic gestures hold immense promise for HCI, static gesture recognition poses a unique set of challenges [15,16]. Unlike dynamic gestures, where the motion itself provides valuable context, static gestures rely solely on the configuration of the hand or fingers. This lack of inherent movement makes static gestures more challenging to classify accurately, as variations in hand orientation and individual anatomical differences can introduce ambiguity. Thus, the quest for robust and accurate static gesture recognition methods becomes crucial in harnessing the full potential of gesture-based HCI.

In this paper, we investigate the capabilities of proprietary Ultra-Wideband (UWB) radar for the meticulous collection of static gesture data. Our objective is to employ this powerful sensor technology to capture, preprocess, and prepare the dataset meticulously. Subsequently, we embark on the task of training high-accuracy Convolutional Neural Network (CNN) models. We aim to address the intricate challenges associated with static gesture recognition, paving the way for a more seamless and natural human-machine interaction paradigm, akin to human-to-human communication.

## 2 LITERATURE REVIEW

Gesture recognition has evolved as a critical component of Human-Computer Interaction (HCI) with the integration of advanced technology into daily life [1-8]. In response to the growing demand for more seamless HCI methods, hand gesture recognition-based HCI has emerged as a promising avenue for enhancing man-machine interactions [1,2].

One notable advancement comes from Ahmed and Cho (2020), who presented a groundbreaking technique utilizing Impulse-Radio Ultra-Wideband (IR-UWB) radar and an inception module-based classifier [1]. This approach achieved a remarkable gesture recognition accuracy of 95% by transforming radar signals into three-dimensional image patterns and analyzing them using the inception module-based variant of GoogLeNet [1]. Their work demonstrates the potential of deep learning in enhancing gesture recognition accuracy.

Building upon these efforts, Li et al. (2021) harnessed the power of IR-UWB radar, employing ShuffleNet V2, a lightweight Convolutional Neural Network (CNN) architecture, to achieve an astounding accuracy of 98.52% [2]. This impressive accuracy, coupled with the conversion of time-domain radar signals into continuous Range-Doppler Maps (RDM), highlights the effectiveness of radar-based gesture recognition in terms of accuracy, speed, and robustness [2].

In the realm of vehicular applications, Khan and Cho (2017) explored gesture recognition within cars using IR-UWB radar [3]. Their work demonstrated the potential for neural networks to recognize distinct gestures made in front of a radar sensor, enabling gesture-based control of various in-car electronics. Meanwhile, Khan et al. (2017) tackled the challenge of reducing visual attention while driving by introducing a real-time hand-based gesture recognition algorithm. Their approach relied on three independent features—variance of the probability density function, time of arrival variation, and frequency of the reflected signal—to classify gestures with robustness against changes in distance or direction, even amidst unrelated motions within the car [4].

Beyond traditional gesture recognition, Li et al. (2021) expanded the scope to include sign language recognition, a crucial tool for effectively communicating with the deaf and mute communities. Their novel discriminative feature, cumulative distribution density (CDD), enabled a significant 8.6% improvement in gesture recognition accuracy, proving its versatility across different types of gestures. This development underscores the potential of radar technology in facilitating inclusive communication [5].

Furthermore, radar technology has found applications in industrial environments, with Delamare et al. (2020) evaluating the performance of UWB localization systems for precise indoor localization [6]. In this context, UWB systems demonstrated the capability to estimate the position of a person moving in complex industrial environments, offering millimetric accuracy when compared to a motion capture system [6].

Li et al. (2019) proposed a hierarchical sensor fusion approach, where radar acted as an enhancer alongside pressure sensor arrays for micro-gesture recognition. The use of sequential forward selection in feature extraction significantly reduced computational complexity while improving classification performance. Additionally, soft and hard fusion methods enhanced classification accuracy and reduced false alarms, highlighting the potential of sensor fusion in radar-based gesture recognition [7].

Moreover, Park et al. (2020) introduced a time-domain-based AI radar system, achieving recognition rates of 93.2% and 90.5% for static and dynamic gestures, respectively [8]. The utilization of high-speed sampling through a time-extension technique allowed AI to process time-domain radar signals, recognizing both static and dynamic gestures with remarkable accuracy [8]. This breakthrough highlights the importance of radar technology in providing a versatile and precise means of gesture recognition for various applications [8].

The evolution of radar-based gesture recognition has demonstrated its potential to revolutionize HCI across different domains, with ongoing advancements enhancing accuracy, feature extraction, and versatility.

## 3 Materials And Methods

### 3.1 Dataset Collection and Characteristics

The dataset employed in this study has undergone meticulous curation to encompass a diverse array of hand gestures executed at varying distances of 10 centimetres, 25 centimetres, and 50 centimetres. This thoughtful stratification ensures the dataset captures a rich spectrum of spatial characteristics, accommodating potential visual variations resulting from differing proximities. Further enhancing its complexity, the dataset includes universally recognizable hand gestures, such as "OK," "VICTORY," "STOP," "PALM," and "LIKE." This comprehensive assortment facilitates the robust training of a gesture recognition model proficient in distinguishing among a myriad of hand gestures.

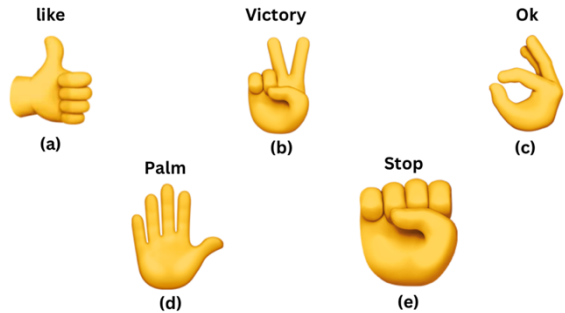

Fig1. Static Gestures

To challenge the recognition model and elevate dataset complexity, we introduced a "no-gesture" class. This addition simulates periods of inactivity or non-gestural activity, demanding the model to exhibit the discernment needed to differentiate authentic gestures from phases of quiescence.

Indeed, the foundational data for this dataset originates from proprietary UWB Radar sensor technology. This advanced sensor forms the backbone of a meticulous data collection process that spans 9360 seconds. The average data collection duration per gesture amounts to 1560 seconds, with data gathered from 5 to 12 subjects for different gestures. This approach ensures both stratification and randomness within the dataset, thereby further enhancing its diversity and richness.

### 3.2 Dataset Preprocessing

Several essential pre-processing operations were performed to ensure the integrity and suitability of the data for subsequent model training and analysis.

**Outlier mitigation**

Robust normalization techniques, such as the rolling median operation and Z-score computation, were used to identify and remove data points with anomalous values that exceeded a predefined threshold. This process significantly improved the quality of the dataset by reducing noise and preserving salient data features.

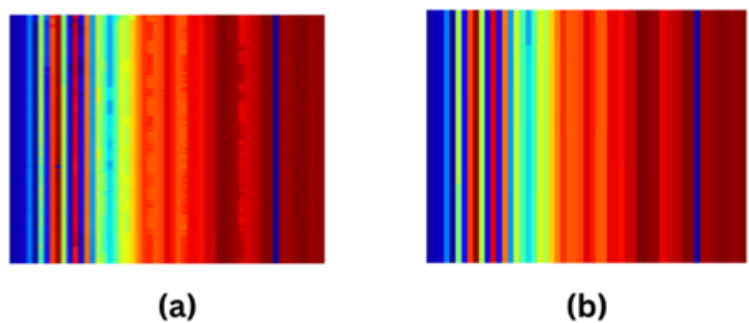

Fig 2. (a) Without Outlier Mitigation (b) With Outlier Mitigation

**False colour image transformation**

A sophisticated transformation process was used to convert the normalized data into false colour images. This approach was strategically selected to enhance the visibility of underlying patterns in the dataset. By translating the data into a visual representation with distinct colour mappings, this process facilitated more discernible pattern recognition and provided an intuitive means of visualizing the gestural data.

**Aspect ratio-preserving resizing**

An aspect ratio-preserving resizing technique was systematically employed to standardize the dimensions of the images derived from the dataset. This ensured that all images conformed to a consistent size format, specifically 159x200 pixels in the RGB A colour space. This standardization was essential for preserving the original structural attributes of the gestures while concurrently rendering the data compatible with subsequent model training processes.

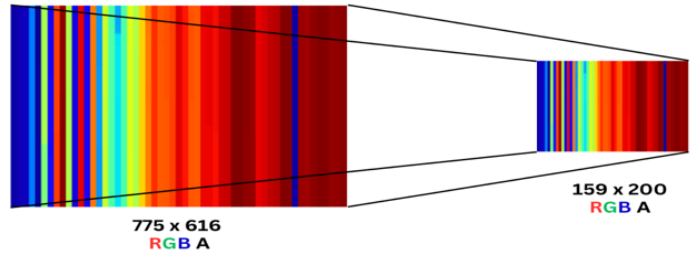

Fig 3. (a) Resizing without losing information

These pivotal pre-processing steps collectively fortified the dataset's suitability for advanced analyses and model development, serving as a critical preparatory phase in the pursuit of accurate and meaningful gesture recognition outcomes.

### 3.3 Model Training

The dataset was randomly partitioned into training, validation, and testing subsets in a 70:15:15 ratio. This standard procedure ensured that the model was trained on a representative sample of the data and evaluated on unseen data.

A variety of deep learning architectures were systematically trained, including different variations of Convolutional Neural Networks (CNNs) and MobileNets. These architectures were chosen for their demonstrated ability to learn spatial features from images, which is essential for gesture recognition.

### 3.4 Model Evaluation Results

In the evaluation of various deep learning models for UWB-based static gesture classification, the following results were obtained.

The models were identified by their respective IDs and categorized into two main types: Convolutional Neural Networks (CNNs) and mobile networks. The evaluation metrics included both subcategory accuracy (SubClass-Accuracy %) and supercategory accuracy (SuperClass-Accuracy %).

CNN-based models outperformed MobileNet models on both subcategory and supercategory accuracy. CNN v3 achieved the highest overall performance, followed by CNN v2. Among the MobileNet models, Mbnet V3 and Mbnet V5 achieved the best results, but their accuracy was generally lower than that of the CNN models.

These findings (See Table 1.) suggest that CNN-based architectures are better suited for UWB-based static gesture classification.

**Table 1. Accuracy of different models**

| Si.No | Model ID | Model Type | Sub Class Accuracy % | Super Class Accuracy % |
|---|---|---|---|---|
| 1 | CNN v3 | CNN | 95.76% | 96.78% |
| 2 | CNN v1 | CNN | 92.47% | 91.14% |
| 3 | MobileNet v5 | MobileNet | 88.14% | 91.75% |
| 4 | CNN v2 | CNN | 88.14% | 91.75% |
| 5 | MobileNet v3 | MobileNet | 84.87% | 89.78% |
| 6 | MobileNet v1 | MobileNet | 80.58% | 83.22% |
| 7 | MobileNet v2 | MobileNet | 74.55% | 79.03% |
| 8 | MobileNet v6 | MobileNet | 76.14% | 72.10% |

The model evaluation was performed on an Apple MacBook Pro 2019 with 8GB of RAM and using the CPU.

Two critical metrics for evaluating model performance are Inference Time and Process Time.

- Inference Time: The time it takes a model to make predictions on input data. This is a measure of the model's responsiveness and speed.

- Process Time: The total time it takes to process input data and obtain predictions, including Inference Time and any additional time required for data preprocessing. This provides a comprehensive view of the end-to-end processing efficiency of the gesture recognition system.

Memory allocation is managed based on the unique process ID associated with the execution of the model inference task. This ensures that the model has access to the requisite memory resources for its computations.

In our empirical findings, we observed that CNN models had an average memory usage of approximately 746 MB, while MobileNets had an average memory usage of approximately 450 MB. Both model architectures achieved commendable process times, each completing their tasks in less than a second (> 1 Second).

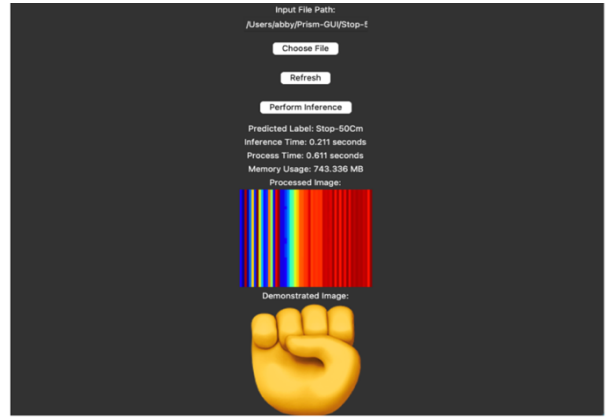

Fig 4. GUI To Simulate Real-Life Gesture Recognition.

These results suggest that both CNNs and Mobile Nets are efficient for processing and recognizing hand gestures. Mobile Nets have a lower memory footprint, but CNNs are slightly faster. This information can be used to optimize resource utilization in real-world applications.

## 4      Results And Discussions

This study presents a meticulously curated dataset of hand gestures captured at various distances using UWB Radar sensor technology. The dataset encompasses diverse gestures, including "OK," "VICTORY," "STOP," "PALM," and "LIKE," along with a "no-gesture" class for added complexity. Essential pre-processing steps, such as outlier mitigation, false-colour image transformation, and aspect ratio-preserving resizing, enhance data quality. Model training involved CNN and MobileNet architectures, with CNN v3 achieving the highest accuracy at 96.78%. Empirical findings indicate efficient memory usage and fast processing times for both model types. CNNs exhibit a slightly higher memory footprint but offer faster performance, while MobileNets are memory-efficient. These insights can inform resource optimization in practical applications.

**DATA AVAILABILITY**

The author Doesn't have the right to share the Dataset, the Preprocessing and Model Code can be provided on request.